\documentclass[conference]{IEEEtran}
\usepackage{cite}
\usepackage{amsmath,amssymb,amsfonts}
\usepackage{graphicx}
\usepackage{textcomp}
\usepackage{xcolor}
\usepackage{booktabs}
\usepackage{multirow}
\usepackage{url}
\usepackage{algorithm}
\usepackage{algpseudocode}

\def\BibTeX{{\rm B\kern-.05em{\sc i\kern-.025em b}\kern-.08em
    T\kern-.1667em\lower.7ex\hbox{E}\kern-.125emX}}

\begin{document}

\title{Intuitionistic Fuzzy Sets for Large Language Model Data Annotation: A Novel Approach to Side-by-Side Preference Labeling}

\author{\IEEEauthorblockN{Yimin Du}
\IEEEauthorblockA{\textit{BeiJing} \\
sa613403@mail.ustc.edu.cn}}

\maketitle

\begin{abstract}
The quality of human preference data is crucial for training and evaluating large language models (LLMs), particularly in reinforcement learning from human feedback (RLHF) and direct preference optimization (DPO) scenarios. Traditional side-by-side (SBS) annotation approaches often struggle with inherent uncertainty, annotator disagreement, and the complexity of preference judgments. This paper introduces a novel framework based on intuitionistic fuzzy sets (IFS) for modeling and aggregating human preferences in LLM data annotation tasks. Our approach captures not only the degree of preference but also the uncertainty and hesitation inherent in human judgment through membership, non-membership, and hesitation degrees. We propose an IFS-based annotation protocol that enables more nuanced preference modeling, develops aggregation methods for handling annotator disagreement, and introduces quality metrics for preference data assessment. Experimental validation on multiple datasets demonstrates that our IFS-based approach significantly improves annotation consistency, reduces annotator fatigue, and produces higher-quality preference data compared to traditional binary and Likert-scale methods. The resulting preference datasets lead to improved model performance in downstream tasks, with 12.3\% improvement in win-rate against baseline models and 15.7\% reduction in annotation time. Our framework provides a principled approach to handling uncertainty in human preference annotation and offers practical benefits for large-scale LLM training.
\end{abstract}


\section{Introduction}

The rapid advancement of large language models (LLMs) has fundamentally transformed natural language processing, with models like GPT-4, Claude, and Gemini demonstrating remarkable capabilities across diverse tasks. However, the effectiveness of these models heavily depends on high-quality training data, particularly human preference data used in reinforcement learning from human feedback (RLHF) \cite{ouyang2022training} and direct preference optimization (DPO) \cite{rafailov2024direct}. The quality of preference annotations directly impacts model alignment, safety, and overall performance.

Traditional approaches to preference annotation typically employ binary choices or Likert scales in side-by-side (SBS) evaluation scenarios, where annotators compare two or more model outputs and indicate their preference. However, these methods face several fundamental challenges:

\textbf{Inherent Uncertainty:} Human preferences often involve uncertainty, especially when comparing outputs of similar quality or when dealing with subjective criteria. Traditional binary annotation fails to capture this uncertainty, forcing annotators to make definitive choices even when they are genuinely uncertain.

\textbf{Annotator Disagreement:} Different annotators may have varying preferences based on their expertise, cultural background, or interpretation of evaluation criteria. Current aggregation methods often oversimplify this disagreement through majority voting or averaging, potentially losing valuable information about the diversity of human judgment.

\textbf{Complexity of Preference Judgments:} Real-world preference decisions involve multiple criteria (fluency, accuracy, helpfulness, safety) that may conflict with each other. Traditional annotation schemes struggle to capture the multifaceted nature of these judgments.

\textbf{Annotation Fatigue:} The cognitive burden of making definitive preference decisions, especially for subtle differences, can lead to annotator fatigue and reduced annotation quality over time.

To address these challenges, we propose a novel framework based on intuitionistic fuzzy sets (IFS) \cite{atanassov1986intuitionistic} for modeling human preferences in LLM data annotation. IFS extends classical fuzzy sets by introducing three fundamental components: membership degree (support for a preference), non-membership degree (opposition to a preference), and hesitation degree (uncertainty or indecision). This mathematical framework naturally captures the complexity and uncertainty inherent in human preference judgments.

Our main contributions are:

\textbf{IFS-Based Annotation Framework:} We develop a comprehensive annotation protocol that allows annotators to express preferences using membership, non-membership, and hesitation degrees, providing a more nuanced representation of human judgment.

\textbf{Aggregation Methods for IFS Preferences:} We propose novel aggregation techniques specifically designed for combining IFS-based preference annotations from multiple annotators, preserving uncertainty information while producing reliable consensus judgments.

\textbf{Quality Assessment Metrics:} We introduce IFS-specific quality metrics for evaluating annotation consistency, uncertainty levels, and overall data quality, enabling better quality control in large-scale annotation projects.

\textbf{Empirical Validation:} We conduct extensive experiments on multiple datasets and tasks, demonstrating that our IFS-based approach improves annotation quality, reduces annotator fatigue, and leads to better downstream model performance.

\textbf{Practical Implementation:} We provide a complete implementation framework including annotation interfaces, aggregation algorithms, and integration methods for existing LLM training pipelines.

The remainder of this paper is organized as follows: Section II reviews related work in preference annotation and fuzzy set applications. Section III introduces the theoretical foundation of our IFS-based approach. Section IV details the annotation framework and aggregation methods. Section V presents experimental validation and results. Section VI discusses practical implications and limitations. Section VII concludes with future directions.

\section{Related Work}

\subsection{Human Preference Annotation for LLMs}

Human preference annotation has become a cornerstone of modern LLM training, particularly following the success of InstructGPT \cite{ouyang2022training}. The standard approach involves presenting annotators with pairs of model outputs and asking them to indicate which response is better according to specific criteria such as helpfulness, harmlessness, and honesty.

Recent work has explored various aspects of preference annotation quality. Kreutzer et al. \cite{kreutzer2018can} investigated the reliability of human judgments in machine translation evaluation, highlighting significant inter-annotator disagreement. Ethayarajh et al. \cite{ethayarajh2022understanding} analyzed the consistency of human preferences in text generation tasks, finding that annotator agreement varies significantly across different types of tasks and criteria.

Several studies have attempted to improve annotation quality through better interface design \cite{clark2021all}, clearer guidelines \cite{lambert2023rewardbench}, and annotator training \cite{gordon2021jury}. However, these approaches primarily focus on improving the annotation process rather than fundamentally addressing the uncertainty inherent in human judgment.

\subsection{Uncertainty in Human Annotation}

The challenge of uncertainty in human annotation has been recognized across various NLP tasks. Plank et al. \cite{plank2014learning} demonstrated that annotator disagreement often contains valuable signal rather than noise, suggesting that preserving uncertainty information can be beneficial for model training.

In the context of preference annotation, recent work has begun to explore alternatives to binary choices. Dubois et al. \cite{dubois2024alpacafarm} introduced multi-way comparisons and tie options, while Bai et al. \cite{bai2022constitutional} explored constitutional AI approaches that incorporate uncertainty in preference modeling.

However, existing approaches to handling uncertainty remain limited, typically involving simple extensions like "tie" options or confidence scores, without providing a principled mathematical framework for uncertainty representation.

\subsection{Fuzzy Sets in Natural Language Processing}

Fuzzy set theory, introduced by Zadeh \cite{zadeh1965fuzzy}, has been applied to various NLP tasks where uncertainty and gradual membership are important. Applications include sentiment analysis \cite{zhang2018fuzzy}, text classification \cite{li2019fuzzy}, and information retrieval \cite{bordogna2002fuzzy}.

Intuitionistic fuzzy sets, proposed by Atanassov \cite{atanassov1986intuitionistic}, extend classical fuzzy sets by explicitly modeling both membership and non-membership degrees, along with a hesitation degree representing uncertainty. IFS has been successfully applied to decision-making problems \cite{xu2007intuitionistic}, multi-criteria evaluation \cite{chen2011fuzzy}, and group decision support systems \cite{szmidt2005using}.

Despite the natural fit between IFS and preference modeling, there has been limited exploration of IFS applications in LLM annotation tasks. Our work fills this gap by developing a comprehensive IFS-based framework specifically designed for preference annotation in LLM training.

\subsection{Aggregation Methods for Fuzzy Preferences}

The aggregation of fuzzy preferences has been extensively studied in decision theory and multi-criteria decision making. Xu \cite{xu2007intuitionistic} proposed various aggregation operators for intuitionistic fuzzy information, including weighted averaging and ordered weighted averaging operators.

Yager \cite{yager2009some} developed aggregation functions specifically for intuitionistic fuzzy sets, focusing on preserving the uncertainty information during the aggregation process. More recently, Liu and Wang \cite{liu2018aggregation} introduced dynamic aggregation methods that adapt to the level of consensus among decision makers.

However, existing aggregation methods have not been specifically adapted for the unique challenges of preference annotation in LLM training, where factors such as annotator expertise, task difficulty, and annotation quality must be considered.

\section{Theoretical Foundation}

\subsection{Intuitionistic Fuzzy Sets}

An intuitionistic fuzzy set (IFS) $A$ in a universe $X$ is defined as:

\begin{equation}
A = \{(x, \mu_A(x), \nu_A(x)) | x \in X\}
\end{equation}

where $\mu_A(x) \in [0,1]$ represents the membership degree (degree of belonging), $\nu_A(x) \in [0,1]$ represents the non-membership degree (degree of not belonging), and the constraint $\mu_A(x) + \nu_A(x) \leq 1$ must be satisfied for all $x \in X$.

The hesitation degree is defined as:
\begin{equation}
\pi_A(x) = 1 - \mu_A(x) - \nu_A(x)
\end{equation}

This hesitation degree captures the uncertainty or indecision about the membership of element $x$ in set $A$.

\subsection{IFS for Preference Modeling}

In the context of preference annotation, we model each preference judgment as an IFS where:

- $\mu(x)$: degree of preference for option $x$
- $\nu(x)$: degree of preference against option $x$  
- $\pi(x)$: degree of uncertainty/hesitation about option $x$

For a side-by-side comparison between two responses $R_1$ and $R_2$, an annotator's preference can be represented as:

\begin{equation}
P = \{(R_1, \mu_1, \nu_1), (R_2, \mu_2, \nu_2)\}
\end{equation}

with the constraint that $\mu_1 + \nu_1 \leq 1$ and $\mu_2 + \nu_2 \leq 1$.

\subsection{Properties of IFS Preferences}

Our IFS preference model satisfies several important properties:

\textbf{Completeness:} The sum of membership, non-membership, and hesitation degrees equals 1, ensuring complete representation of the annotator's judgment.

\textbf{Consistency:} The constraint $\mu + \nu \leq 1$ prevents logical contradictions where an annotator simultaneously strongly prefers and strongly opposes the same option.

\textbf{Uncertainty Preservation:} The hesitation degree explicitly captures uncertainty, allowing annotators to express genuine indecision without forcing artificial binary choices.

\textbf{Gradual Preference:} Unlike binary choices, IFS allows for gradual preference expression, better reflecting the continuous nature of human judgment.

\section{IFS-Based Annotation Framework}

\subsection{Annotation Protocol Design}

\subsubsection{Interface Design}

We developed an annotation interface that allows annotators to specify their preferences using three intuitive sliders for each response option:

1. \textbf{Support Slider ($\mu$):} Indicates how much the annotator supports or prefers this response (0-100\%)
2. \textbf{Opposition Slider ($\nu$):} Indicates how much the annotator opposes or dislikes this response (0-100\%)
3. \textbf{Automatic Hesitation Display ($\pi$):} Automatically calculated and displayed as $100\% - \mu - \nu$

The interface includes real-time validation to ensure the constraint $\mu + \nu \leq 100\%$ is maintained, with visual feedback when the constraint is violated.

\subsubsection{Annotation Guidelines}

We developed comprehensive guidelines for IFS-based annotation:

\textbf{Support Degree ($\mu$):} Rate how much you agree with or prefer this response based on the evaluation criteria. High values (70-100\%) indicate strong preference, medium values (30-70\%) indicate moderate preference, and low values (0-30\%) indicate weak or no preference.

\textbf{Opposition Degree ($\nu$):} Rate how much you disagree with or dislike this response. This is not simply the inverse of support - you might have low support for a response without actively opposing it.

\textbf{Hesitation Degree ($\pi$):} This automatically reflects your uncertainty. High hesitation indicates you're unsure about the quality of the response, while low hesitation indicates confidence in your judgment.

\textbf{Practical Examples:}
- Clear preference: $\mu = 0.8, \nu = 0.1, \pi = 0.1$
- Clear rejection: $\mu = 0.1, \nu = 0.8, \pi = 0.1$
- Uncertainty: $\mu = 0.3, \nu = 0.2, \pi = 0.5$
- Neutral: $\mu = 0.4, \nu = 0.4, \pi = 0.2$

\subsection{Multi-Criteria Evaluation}

For complex evaluation scenarios involving multiple criteria (e.g., fluency, accuracy, helpfulness, safety), we extend the IFS framework to handle multi-criteria preferences:

\begin{equation}
P_i = \{(R_j, \mu_{ij}, \nu_{ij}) | j \in \{1,2,...,n\}\}
\end{equation}

where $P_i$ represents the preference for criterion $i$, and $R_j$ represents response option $j$.

The overall preference is computed using weighted aggregation:

\begin{equation}
\mu_{overall}(R_j) = \sum_{i=1}^{m} w_i \cdot \mu_{ij}
\end{equation}

\begin{equation}
\nu_{overall}(R_j) = \sum_{i=1}^{m} w_i \cdot \nu_{ij}
\end{equation}

where $w_i$ represents the weight of criterion $i$ and $\sum_{i=1}^{m} w_i = 1$.

\subsection{Aggregation Methods}

\subsubsection{Individual Annotator Aggregation}

For aggregating multiple annotations from the same annotator across different examples, we use the intuitionistic fuzzy weighted averaging (IFWA) operator:

\begin{equation}
IFWA(A_1, A_2, ..., A_n) = \left(\prod_{i=1}^{n}(1-\mu_i)^{w_i}, 1-\prod_{i=1}^{n}(1-\nu_i)^{w_i}\right)
\end{equation}

where $w_i$ represents the weight of annotation $i$.

\subsubsection{Multi-Annotator Aggregation}

For combining annotations from multiple annotators, we propose a consensus-based aggregation method that considers annotator reliability and agreement levels:

\begin{algorithm}
\caption{Multi-Annotator IFS Aggregation}
\begin{algorithmic}[1]
\State \textbf{Input:} Annotations $\{A_1, A_2, ..., A_k\}$ from $k$ annotators
\State \textbf{Input:} Annotator weights $\{w_1, w_2, ..., w_k\}$
\State \textbf{Output:} Aggregated preference $A_{agg}$

\For{each response option $R_j$}
    \State Compute weighted membership: $\mu_{agg}(R_j) = \sum_{i=1}^{k} w_i \cdot \mu_i(R_j)$
    \State Compute weighted non-membership: $\nu_{agg}(R_j) = \sum_{i=1}^{k} w_i \cdot \nu_i(R_j)$
    \State Normalize if $\mu_{agg}(R_j) + \nu_{agg}(R_j) > 1$:
    \State \quad $\mu_{agg}(R_j) = \frac{\mu_{agg}(R_j)}{\mu_{agg}(R_j) + \nu_{agg}(R_j)}$
    \State \quad $\nu_{agg}(R_j) = \frac{\nu_{agg}(R_j)}{\mu_{agg}(R_j) + \nu_{agg}(R_j)}$
    \State Compute hesitation: $\pi_{agg}(R_j) = 1 - \mu_{agg}(R_j) - \nu_{agg}(R_j)$
\EndFor
\end{algorithmic}
\end{algorithm}

\subsubsection{Dynamic Weight Adjustment}

We introduce a dynamic weight adjustment mechanism based on annotator consistency and expertise:

\begin{equation}
w_i = \alpha \cdot consistency_i + \beta \cdot expertise_i + \gamma \cdot agreement_i
\end{equation}

where:
- $consistency_i$: measured by the variance of hesitation degrees across annotations
- $expertise_i$: based on annotator background and performance on gold standard examples
- $agreement_i$: level of agreement with other annotators
- $\alpha + \beta + \gamma = 1$

\subsection{Quality Metrics}

We define several quality metrics specifically for IFS-based annotations:

\subsubsection{Annotation Confidence}

\begin{equation}
Confidence = 1 - \frac{1}{n}\sum_{i=1}^{n}\pi_i
\end{equation}

Higher confidence indicates lower overall uncertainty in annotations.

\subsubsection{Preference Clarity}

\begin{equation}
Clarity = \frac{1}{n}\sum_{i=1}^{n}|\mu_i - \nu_i|
\end{equation}

Higher clarity indicates more decisive preferences.

\subsubsection{Inter-Annotator Agreement}

For IFS annotations, we extend traditional agreement measures:

\begin{equation}
IFS\_Agreement = 1 - \frac{1}{k(k-1)}\sum_{i=1}^{k}\sum_{j=i+1}^{k}d_{IFS}(A_i, A_j)
\end{equation}

where $d_{IFS}$ is the intuitionistic fuzzy distance:

\begin{equation}
d_{IFS}(A_i, A_j) = \sqrt{\frac{1}{2}[(\mu_i-\mu_j)^2 + (\nu_i-\nu_j)^2 + (\pi_i-\pi_j)^2]}
\end{equation}

\section{Experimental Validation}

\subsection{Experimental Setup}

\subsubsection{Datasets}

We conducted experiments on three diverse datasets:

\textbf{HelpSteer Dataset:} 37,000 response pairs from various helpful AI assistant scenarios, covering question answering, creative writing, and problem-solving tasks.

\textbf{Safety Evaluation Dataset:} 15,000 response pairs focusing on AI safety, including potentially harmful content detection and appropriate response generation.

\textbf{Summarization Dataset:} 25,000 response pairs for document summarization tasks, evaluating factual accuracy, conciseness, and completeness.

\subsubsection{Annotation Setup}

We recruited 15 professional annotators with backgrounds in linguistics, computer science, and domain expertise relevant to each dataset. Annotators were randomly divided into three groups:

- \textbf{Binary Group (5 annotators):} Traditional binary preference annotation
- \textbf{Likert Group (5 annotators):} 5-point Likert scale annotation
- \textbf{IFS Group (5 annotators):} Our proposed IFS-based annotation

Each group annotated the same subset of 1,000 examples from each dataset, with 20\% overlap for inter-group comparison.

\subsubsection{Training and Guidelines}

All annotators received standardized training including:
- 2-hour training session on evaluation criteria
- Practice annotation on 100 examples with feedback
- Detailed written guidelines specific to each annotation method
- Regular calibration sessions throughout the annotation process

\subsection{Annotation Quality Results}

\subsubsection{Inter-Annotator Agreement}

Table \ref{tab:agreement} shows inter-annotator agreement across different annotation methods:

\begin{table}[htbp]
\caption{Inter-Annotator Agreement Comparison}
\begin{center}
\begin{tabular}{|l|c|c|c|}
\hline
\textbf{Dataset} & \textbf{Binary} & \textbf{Likert} & \textbf{IFS} \\
\hline
HelpSteer & 0.67 & 0.72 & \textbf{0.78} \\
Safety & 0.71 & 0.75 & \textbf{0.82} \\
Summarization & 0.64 & 0.69 & \textbf{0.76} \\
\hline
\textbf{Average} & 0.67 & 0.72 & \textbf{0.79} \\
\hline
\end{tabular}
\end{center}
\label{tab:agreement}
\end{table}

The IFS-based approach consistently achieves higher inter-annotator agreement, with an average improvement of 17.9\% over binary annotation and 9.7\% over Likert scale annotation.

\subsubsection{Annotation Consistency}

We measured annotation consistency by having annotators re-annotate 10\% of examples after a one-week interval:

\begin{table}[htbp]
\caption{Annotation Consistency (Self-Agreement)}
\begin{center}
\begin{tabular}{|l|c|c|c|}
\hline
\textbf{Method} & \textbf{Consistency} & \textbf{Std Dev} & \textbf{Improvement} \\
\hline
Binary & 0.73 & 0.12 & - \\
Likert & 0.76 & 0.11 & +4.1\% \\
IFS & \textbf{0.84} & \textbf{0.08} & \textbf{+15.1\%} \\
\hline
\end{tabular}
\end{center}
\label{tab:consistency}
\end{table}

\subsubsection{Annotator Fatigue Analysis}

We tracked annotation time and quality degradation over extended annotation sessions:

\begin{table}[htbp]
\caption{Annotator Fatigue Metrics}
\begin{center}
\begin{tabular}{|l|c|c|c|}
\hline
\textbf{Metric} & \textbf{Binary} & \textbf{Likert} & \textbf{IFS} \\
\hline
Avg. Time per Example (sec) & 45.2 & 52.8 & \textbf{38.1} \\
Quality Degradation Rate & 0.08/hour & 0.06/hour & \textbf{0.03/hour} \\
Reported Difficulty (1-10) & 6.8 & 7.2 & \textbf{5.4} \\
\hline
\end{tabular}
\end{center}
\label{tab:fatigue}
\end{table}

The IFS approach reduces annotation time by 15.7\% compared to binary annotation and shows significantly lower quality degradation over time.

\subsection{Downstream Model Performance}

\subsubsection{Preference Model Training}

We trained preference models using data from each annotation method and evaluated their performance on held-out test sets:

\begin{table}[htbp]
\caption{Preference Model Performance}
\begin{center}
\begin{tabular}{|l|c|c|c|c|}
\hline
\textbf{Training Data} & \textbf{Accuracy} & \textbf{F1-Score} & \textbf{AUC} & \textbf{Calibration} \\
\hline
Binary Annotations & 0.742 & 0.738 & 0.801 & 0.156 \\
Likert Annotations & 0.758 & 0.751 & 0.823 & 0.142 \\
IFS Annotations & \textbf{0.789} & \textbf{0.784} & \textbf{0.856} & \textbf{0.098} \\
\hline
\end{tabular}
\end{center}
\label{tab:preference_model}
\end{table}

\subsubsection{RLHF Training Results}

We conducted RLHF training using preference models trained on different annotation types:

\begin{table}[htbp]
\caption{RLHF Training Results}
\label{tab:rlhf_results}
\centering
\begin{tabular}{|l|c|c|c|}
\hline
\textbf{Preference} & \textbf{Win Rate} & \textbf{Safety} & \textbf{Helpfulness} \\
\textbf{Data} & \textbf{vs Baseline} & \textbf{Score} & \textbf{Score} \\
\hline
Binary & 0.623 & 0.847 & 0.756 \\
Likert & 0.651 & 0.863 & 0.772 \\
IFS & \textbf{0.699} & \textbf{0.891} & \textbf{0.823} \\
\hline
\end{tabular}
\end{table}

The model trained with IFS-based preference data achieves a 12.3\% improvement in win rate compared to the binary baseline.

\subsection{Uncertainty Analysis}

\subsubsection{Hesitation Patterns}

We analyzed the distribution of hesitation degrees across different types of examples.

The analysis reveals that hesitation degrees correlate strongly with task difficulty and ambiguity, providing valuable insights into annotation reliability.

\subsubsection{Uncertainty-Quality Correlation}

We found a strong negative correlation (-0.73) between average hesitation degree and annotation quality, suggesting that uncertainty information can be used as a quality indicator:

\begin{equation}
Quality\_Score = \alpha \cdot (1 - \pi_{avg}) + \beta \cdot Clarity + \gamma \cdot Consistency
\end{equation}

\subsection{Ablation Studies}

\subsubsection{Aggregation Method Comparison}

We compared different aggregation methods for combining IFS annotations:

\begin{table}[htbp]
\caption{Aggregation Method Comparison}
\begin{center}
\begin{tabular}{|l|c|c|c|}
\hline
\textbf{Method} & \textbf{Agreement} & \textbf{Quality} & \textbf{Efficiency} \\
\hline
Simple Average & 0.74 & 0.78 & 0.95 \\
Weighted Average & 0.76 & 0.81 & 0.87 \\
IFWA Operator & 0.78 & 0.84 & 0.82 \\
Dynamic Weighting & \textbf{0.82} & \textbf{0.87} & 0.79 \\
\hline
\end{tabular}
\end{center}
\label{tab:aggregation}
\end{table}

\subsubsection{Interface Design Impact}

We tested different interface designs for IFS annotation:

\textbf{Slider Interface:} Three separate sliders for $\mu$, $\nu$, and $\pi$ (our main approach)
\textbf{Dial Interface:} Circular dial with three segments
\textbf{Matrix Interface:} 2D grid for selecting $\mu$ and $\nu$ values

The slider interface achieved the best balance of usability and annotation quality.

\section{Discussion}

\subsection{Advantages of IFS-Based Annotation}

Our experimental results demonstrate several key advantages of the IFS-based approach:

\subsubsection{Enhanced Expressiveness}

The IFS framework allows annotators to express nuanced preferences that better reflect the complexity of human judgment. The ability to separately specify support, opposition, and uncertainty provides a more complete picture of annotator sentiment.

\subsubsection{Improved Data Quality}

Higher inter-annotator agreement and consistency suggest that IFS annotations capture more reliable preference information. The explicit modeling of uncertainty helps identify ambiguous cases that may require additional review or expert annotation.

\subsubsection{Reduced Cognitive Load}

Paradoxically, despite providing more annotation dimensions, the IFS approach reduces cognitive load by allowing annotators to express uncertainty rather than forcing difficult binary decisions. This leads to faster annotation and reduced fatigue.

\subsubsection{Better Downstream Performance}

Models trained on IFS-based preference data consistently outperform those trained on traditional annotation methods, suggesting that the additional information captured by IFS translates to improved model capabilities.

\subsection{Practical Implementation Considerations}

\subsubsection{Annotator Training}

While the IFS framework is intuitive, annotators require training to effectively use the three-dimensional preference space. We found that 2-3 hours of training with practice examples is sufficient for most annotators to become proficient.

\subsubsection{Quality Control}

The uncertainty information provided by IFS enables more sophisticated quality control mechanisms. High hesitation degrees can trigger automatic review processes, while low hesitation with high disagreement may indicate annotator bias or guideline ambiguity.

\subsubsection{Computational Overhead}

The additional complexity of IFS annotations requires more sophisticated aggregation algorithms and storage requirements. However, the computational overhead is minimal compared to the benefits in data quality.

\subsection{Limitations and Challenges}

\subsubsection{Complexity for Simple Tasks}

For very simple preference tasks with clear correct answers, the additional complexity of IFS may be unnecessary. Binary annotation may be sufficient and more efficient for such cases.

\subsubsection{Cultural and Individual Differences}

Different annotators may interpret the three dimensions of IFS differently based on their cultural background or personal annotation style. Standardized training and calibration are essential to minimize these differences.

\subsubsection{Integration with Existing Systems}

Integrating IFS-based annotations with existing RLHF and DPO pipelines requires modifications to preference model architectures and training procedures. While these modifications are straightforward, they require additional development effort.

\subsection{Future Directions}

\subsubsection{Adaptive Annotation}

Future work could explore adaptive annotation systems that automatically switch between binary, Likert, and IFS annotation based on task difficulty and annotator confidence.

\subsubsection{Multi-Modal Preferences}

The IFS framework could be extended to handle multi-modal preferences involving text, images, and audio, providing a unified approach to preference modeling across different modalities.

\subsubsection{Active Learning Integration}

The uncertainty information provided by IFS could be leveraged for active learning, prioritizing examples with high hesitation degrees for additional annotation or expert review.

\subsubsection{Automated Quality Assessment}

Machine learning models could be trained to predict annotation quality based on IFS patterns, enabling automated quality control and annotator feedback.

\section{Conclusion}

This paper introduces a novel framework based on intuitionistic fuzzy sets for modeling human preferences in large language model data annotation. Our approach addresses fundamental limitations of traditional binary and Likert-scale annotation methods by explicitly capturing uncertainty and providing more nuanced preference representation.

The key contributions of our work include:

\textbf{Theoretical Foundation:} We provide a mathematically principled approach to preference modeling using IFS, with formal definitions and properties tailored to annotation tasks.

\textbf{Practical Framework:} We develop a complete annotation framework including interface design, guidelines, aggregation methods, and quality metrics specifically designed for IFS-based preference annotation.

\textbf{Empirical Validation:} Extensive experiments demonstrate significant improvements in annotation quality, consistency, and efficiency, with 17.9\% improvement in inter-annotator agreement and 15.7\% reduction in annotation time.

\textbf{Downstream Benefits:} Models trained on IFS-based preference data achieve 12.3\% improvement in win rate compared to traditional annotation methods, demonstrating the practical value of our approach.

Our work opens new directions for preference modeling in AI systems and provides a foundation for more sophisticated approaches to human feedback collection. The explicit modeling of uncertainty not only improves annotation quality but also provides valuable insights into the reliability and difficulty of different annotation tasks.

As large language models continue to evolve and require increasingly sophisticated alignment techniques, frameworks like ours that capture the full complexity of human judgment will become essential for developing safe, helpful, and honest AI systems. The open-source release of our annotation tools and aggregation algorithms will facilitate adoption and further research in this important area.

Future work will focus on extending the framework to multi-modal scenarios, developing adaptive annotation systems, and exploring the integration of IFS-based preferences with advanced training techniques like constitutional AI and self-supervised preference learning.

\section*{Acknowledgment}

We thank the anonymous reviewers for their valuable feedback and suggestions. We acknowledge the professional annotators who participated in our experiments and provided insights into the practical aspects of IFS-based annotation. Special thanks to the fuzzy systems research community for providing the theoretical foundation that made this work possible.

\end{document}